\documentclass[10pt,twocolumn,letterpaper]{article}

\usepackage{cvpr}              

\usepackage{amsmath} 
\usepackage{amssymb}  
\usepackage{multirow}
\usepackage{booktabs}
\usepackage{xcolor}
\usepackage{tabularx}
\usepackage{arydshln}
\newcolumntype{Y}{>{\centering\arraybackslash}X}
\usepackage{makecell}
\usepackage{subcaption}
\usepackage{tcolorbox}
\tcbuselibrary{breakable}
\usepackage{appendix}

\AtBeginEnvironment{appendices}{%
  \crefname{appendix}{Appendix}{Appendices}
  \crefalias{section}{appendix}
  \crefalias{subsection}{appendix}
}

\usepackage{amsmath,amsfonts,bm}

\def\eqref#1{equation~\ref{#1}}

\def\1{\bm{1}}

\DeclareMathAlphabet{\mathsfit}{\encodingdefault}{\sfdefault}{m}{sl}
\SetMathAlphabet{\mathsfit}{bold}{\encodingdefault}{\sfdefault}{bx}{n}

\definecolor{cvprblue}{rgb}{0.21,0.49,0.74}
\usepackage[pagebackref,breaklinks,colorlinks,allcolors=cvprblue]{hyperref}

\def\paperID{5} 
\def\confName{CVPR}
\def\confYear{2026}

\title{Traffic Scene Generation from Natural Language Description for Autonomous Vehicles with Large Language Model}

\author{
Bo-Kai Ruan$^{1}$ \quad Hao-Tang Tsui$^{2}$ \quad Yung-Hui Li$^{3}$ \quad Hong-Han Shuai$^{1}$ \\
$^{1}$National Yang Ming Chiao Tung University\\
$^{2}$Carnegie Mellon University\\
$^{3}$AI Research Center, Hon Hai Research Institute\\
{\tt \small $^{1}$bkruan.ee11@nycu.edu.tw \quad $^{2}$henrytsu@andrew.cmu.edu} \\\\
Project Page: \url{https://basiclab.github.io/TTSG}
\vspace{-8pt}
}

\begin{document}
\maketitle

\begin{abstract}
Generating realistic and controllable traffic scenes from natural language can greatly enhance the development and evaluation of autonomous driving systems. However, this task poses unique challenges: (1) grounding free-form text into spatially valid and semantically coherent layouts, (2) composing scenarios without predefined locations, and (3) planning multi-agent behaviors and selecting roads that respect agents' configurations. To address these, we propose a modular framework, TTSG, comprising prompt analysis, road retrieval, agent planning, and a novel plan-aware road ranking algorithm. While large language models (LLMs) are used as general planners, our design integrates them into a tightly controlled pipeline that enforces structure, feasibility, and scene diversity. Notably, our ranking strategy ensures consistency between agent actions and road geometry, enabling scene generation without predefined routes or spawn points. The framework supports both routine and safety-critical scenarios, as well as multi-stage event composition. Experiments on SafeBench demonstrate that our method achieves the lowest average collision rate (3.5\%) across three critical scenarios. Moreover, driving captioning models trained on our generated scenes improve action reasoning by over 30 CIDEr points. These results underscore our proposed framework for flexible, interpretable, and safety-oriented simulation. 
\end{abstract}
\section{Introduction}
Traffic scene datasets such as nuScenes~\cite{caesar2020nuscenes} and Waymo~\cite{Ettinger_2021_ICCV} have provided rich multimodal driving logs that enable models to understand diverse road scenarios. However, real-world data collection faces inherent limitations due to safety concerns and limited controllability. Therefore, CARLA~\cite{dosovitskiy2017carla} and MetaDrive~\cite{li2021metadrive} have become essential tools for addressing the limitations, particularly for customized scenarios and high-risk events such as aggressive lane changes, obstructed intersections, or near-collision scenarios. These platforms provide safe, scalable, and repeatable environments that enable extensive experimentation across a wide range of driving conditions, including dangerous interactions that are difficult or unsafe to collect in real-world settings. Despite these advantages, scenario generation in existing simulators typically relies on either randomized sampling or replay of logged trajectories. While random sampling helps ensure scenario diversity, it lacks the intentional control necessary to systematically evaluate specific failure modes, edge cases, or model vulnerabilities. On the other hand, replay-based approaches maintain realism but are inherently limited by the distribution of the collected data, making it challenging to generate novel or unseen scenarios, which restricts their general effectiveness.

\vspace{-1pt}

Recent research has explored instruction-driven simulation, where user-defined prompts guide scenario generation. These approaches enhance controllability and enable targeted scene creation. However, several core limitations remain unaddressed. First, methods such as LCTGen~\cite{tan2023language}, ProSim~\cite{tan2024promptable}, and CTG++~\cite{zhong2023ctg} focus primarily on downstream tasks like trajectory prediction or policy learning, and rely on structured or well-specified inputs. As a result, they struggle to handle free-form or ambiguous natural language and lack the capability to construct detailed scene layouts from high-level descriptions. Second, while ChatScene~\cite{zhang2024chatscene} adopts natural language to generate critical scenarios, it focuses solely on agent planning and still requires users to manually specify spawn points and initial map locations. Third, all prior work typically overlooks environmental conditions such as traffic signals, static road objects, and weather, which are essential for creating realistic and diverse driving scenarios. These limitations motivate our modular framework, which directly interprets free-form language descriptions to perform both layout generation and agent planning without relying on rigid prompt formats or predefined spawn points, and further supports conditional generation based on road signals and objects.

Specifically, this paper addresses the upstream task of generating realistic traffic scenarios directly from unstructured natural language descriptions, eliminating the need for predefined maps or manually specified spawn points. To systematically overcome the aforementioned challenges, we propose a training-free, modular framework. First, we leverage a large language model (LLM) to ground the input prompt into explicit and structured scenario elements, including traffic signals, road objects, and weather conditions. Second, based on these parsed conditions, relevant candidate roads are retrieved from a pre-built road graph, and detailed multi-agent behaviors, \eg, agent types, actions, and relative positions, are planned by the LLM. Third, to ensure spatial validity and semantic coherence, we introduce a novel road ranking algorithm. This method evaluates and scores candidate roads according to their alignment with both environmental constraints and planned agent maneuvers, enabling automatic selection of the optimal road layouts. Finally, these scenario components are rendered into executable, realistic traffic scenes through our custom simulation module. By explicitly tackling the challenges of ambiguous prompts, multi-agent planning, and conditional scenario composition, our framework provides an interpretable solution for generating diverse, controllable traffic scenarios purely from natural language.

Our contributions can be summarized as follows:
\begin{itemize}
    \item We propose a novel framework that enables traffic scene generation directly from natural language descriptions, eliminating the need for predefined routes or spawn points, and supporting sequential event simulation.
    \item We introduce a road ranking strategy that jointly considers road conditions and agent plans, ensuring that the layout aligns closely with the input prompt and supports an effective traffic scenario.
    \item Our approach provides trained agents to reduce collision rates for safe driving, and also improves captioning models to narrate and reason about the agent's behaviors.
\end{itemize}

\section{Related Work}

\subsection{Scene Generation with LMs}
Scene generation has attracted significant attention in recent research to improve autonomous driving decisions. Approaches such as~\cite{feng2023trafficgen,zhong2023ctg,ding2025semantically} leverage learning-based methods to schedule agent trajectories on given scenario maps, and \cite{ding2024realgen} incorporates RAG to enhance scene generation by retrieving prior scenarios as guidance. To further integrate language as a controller, LLMs have been introduced as planners or decision-makers. For example, LCTGen~\cite{tan2023language} employs an LLM to convert prompts into structured representations for map retrieval. CTG++~\cite{zhong2023language} uses an LLM to generate code-based losses that guide a diffusion model. ProSim~\cite{tan2024promptable} uses an LM backbone to predict policies for controlling agents. Another line of work focuses on critical scenario generation to simulate high-risk driving events. For instance, ChatScene~\cite{zhang2024chatscene} enables text-to-critical-scenario conversion using an LLM combined with the Scenic language~\cite{fremont2023scenic} to coordinate autonomous agents. In contrast to these methods, we propose a training-free pipeline that performs detailed scene layout and agent planning directly from natural language, enabling flexible environment control without predefined routes or spawn points, and supporting sequential-event design.

\subsection{Autonomous Driving with LMs}

Autonomous driving demands detailed reasoning and planning, motivating the integration of language instructions and reasoning capabilities. For example, \cite{wu2025language} and \cite{wang2025omnidrive} introduce language-based datasets to extend perception modalities from images and LiDAR to text. Given the strength of LMs in handling such information, they have been incorporated into various driving pipelines. ADAPT~\cite{jin2023adapt} utilizes LMs for video captioning and driving decision reasoning. For action planning, approaches such as \cite{chen2024driving,renz2025simlingo,zou2025few} employ LLMs to dynamically plan vehicle actions. To increase safety under critical scenario, LLM-Attacker~\cite{mei2025llm} adopts LLMs to select optimal attacker to train an aggent. In multi-agent systems, \cite{nguyen2024text} uses LLMs to guide the agent to align a target policy, and \cite{gao2025langcoop} explores agent interactions through LLMs. As these works show that LLM agents are suitable for reasoning and planning for driving scenarios, we aim to combine the LLM to generate the traffic scene from text inputs.

\begin{figure*}[t]
    \centering
    \includegraphics[width=\linewidth]{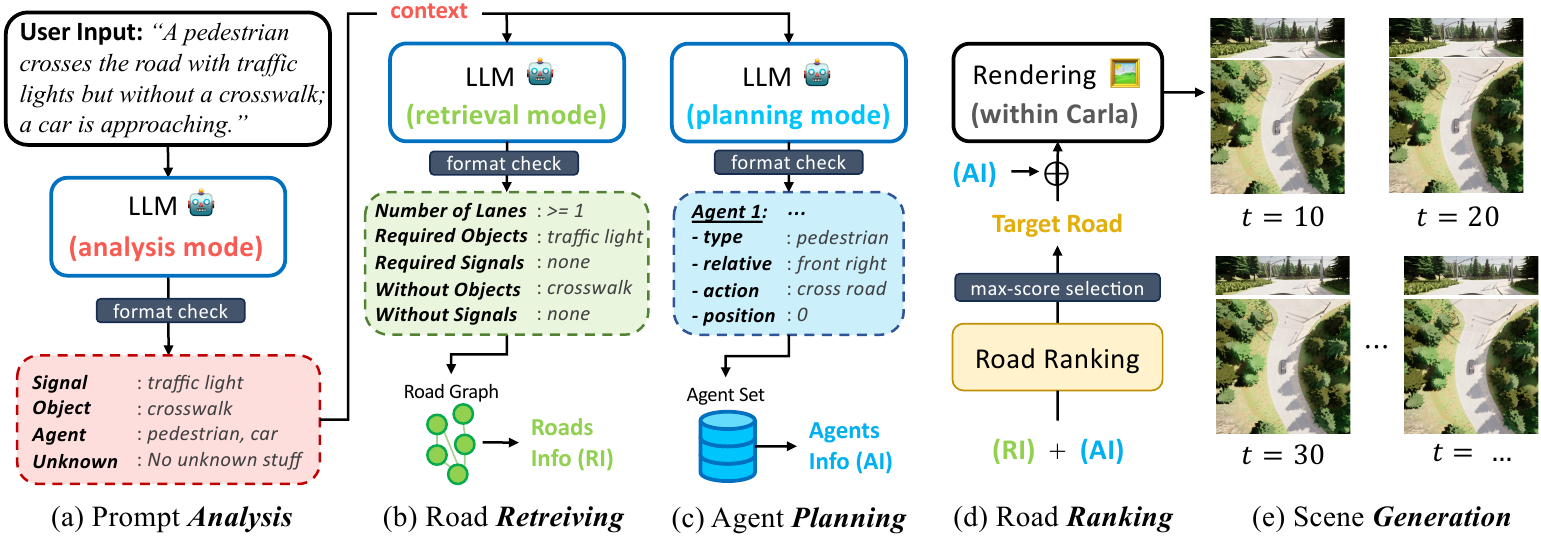}
    \caption{Illustration of the TTSG pipeline, detailing five principal stages: \textit{\textbf{analysis}}, \textit{\textbf{road candidate retrieval}}, \textit{\textbf{agent planning}}, \textit{\textbf{road ranking}}, and \textit{\textbf{generation}}. In the analysis stage, the user’s input is decomposed to provide scene context for subsequent phases. The LLM then determines retrieval conditions to identify appropriate road candidates and facilitates precise planning for agents. When multiple road options are retrieved, a ranking strategy prioritizes roads that best match the agent configurations. Finally, the traffic scene is generated through our custom rendering interface from the selected road and agent information.}
    \label{fig:pipeline}
    \vspace{-10pt}
\end{figure*}

\section{Text-to-Traffic Scene Generation}

In this section, we present our Text-to-Traffic Scene Generation (TTSG) framework. Our approach begins with the construction of a road graph and agent set that categorizes road attributes and agent configurations (\cref{sec:construction}). Given a scene description, as illustrated in~\cref{fig:pipeline}a, the LLM first performs prompt analysis to decompose the input into explicit road conditions and agent requirements (\cref{sec:llm}). This step enables fine-grained control over environment elements as well as detailed agent behaviors and types. Based on the decomposed conditions, relevant road candidates are retrieved from the database and then ranked using a plan-aware strategy that evaluates the compatibility between roads and agent plans. The final selected roads and agent plans are then transformed into a complete traffic scene using our custom rendering interface. We utilize the CARLA simulator as our primary demonstration platform to validate the effectiveness of our pipeline. Various applications and use cases of TTSG are discussed in~\cref{sec:application}.

\subsection{Construction of Road Graph and Agent Set}\label{sec:construction}

To enable automatic spawn point selection, we first construct a road graph $\mathcal{G}$ that encodes detailed road information, which allows us to apply filters to fetch roads that meet specific conditions, promoting automatic scene generation without requiring predefined geometries or spawn positions. This enhances the flexibility of our framework across diverse traffic scenarios. Specifically, we convert each built-in CARLA map into the OpenDRIVE~\cite{asam_opendrive} format, a standardized representation of road networks. We then parse road features such as traffic signals, static objects, junctions, and lane configurations, and organize them into a graph where edges represent road connectivity. This graph structure provides efficient querying of a road and its neighbors, enabling us to determine relationships such as turnability and road linkage. For the agent set, agents are categorized by type, such as regular vehicles, emergency vehicles, or pedestrians, supporting us to fetch supported types and randomly select instances from each category for scenario generation.

\subsection{Retrieval and Planning with LLM}\label{sec:llm}

With the road graph and agent set in place, we next design a mechanism to perform retrieval and planning based on natural language prompts. As illustrated in~\cref{fig:pipeline}, our framework decomposes this process into five modular stages: analysis, retrieval, planning, ranking, and generation. To ensure robust and error-free execution, we apply format verification after each stage, checking that all keys, types, and values are valid and consistent with the expected schema. If any discrepancies are detected, the original prompt is resubmitted to the LLM along with a diagnostic message for automatic correction.  While this iterative process may occasionally introduce additional computational overhead, our results in~\cref{sec:failure} demonstrate that both closed-source and open-source models exhibit very low failure rates.

\begin{figure*}[t]
    \centering
    \begin{subfigure}{\textwidth}
        \centering
        \includegraphics[width=\linewidth]{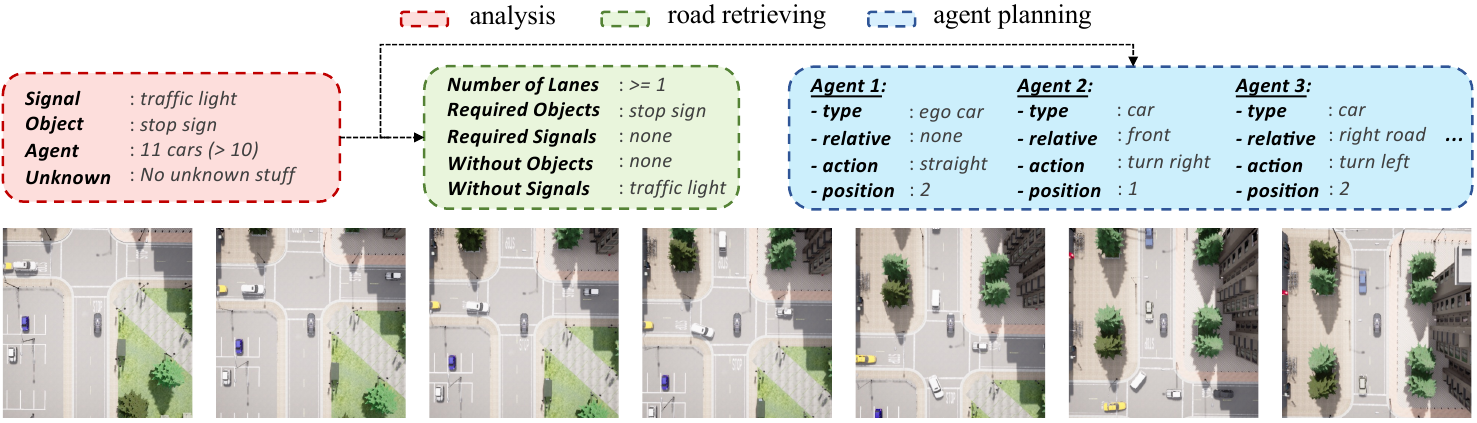}
        \vspace{-15pt}
        \caption{``\textit{Daily traffic at the intersection with more than ten cars, no traffic lights}.''}
        \label{fig:main_normal}
    \end{subfigure}

    \text{}

    \begin{subfigure}{\textwidth}
        \centering
        \includegraphics[width=\linewidth]{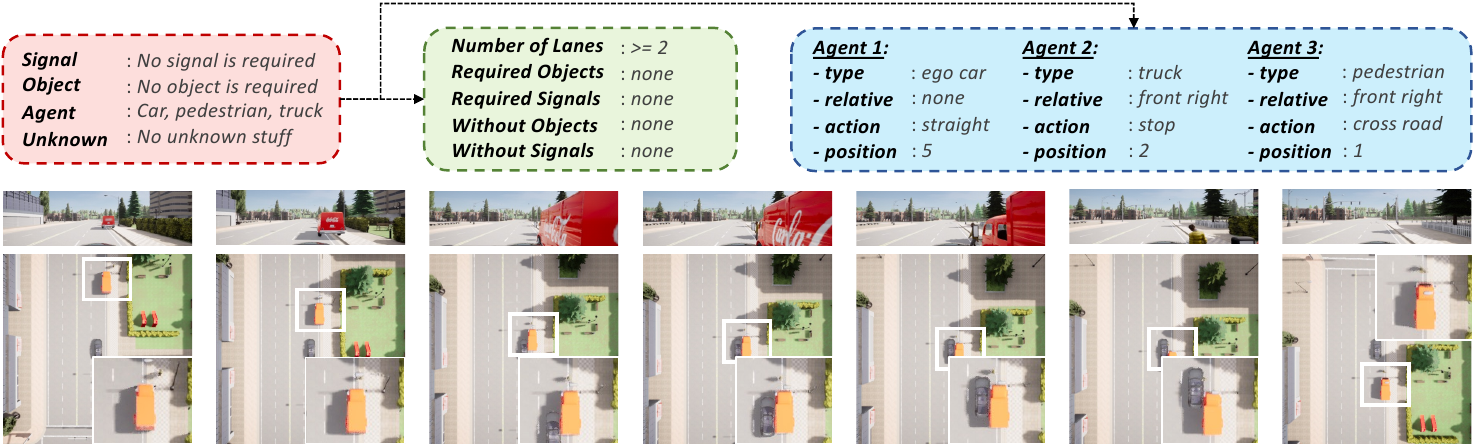}
        \vspace{-15pt}
        \caption{``\textit{A pedestrian on the sidewalk is crossing the street in front of a truck stopping on the shoulder. Both are located on the front right}.''}
        \label{fig:main_critical}
    \end{subfigure}
    \vspace{-15pt}
    \caption{Demonstrations of \textbf{(a)} multi-agent planning, \textbf{(b)} critical scenario with prompts detailed in each sub-caption. Simplified step-by-step outputs from the LLM are shown at the top.}
    \label{fig:scenario}
    \vspace{-8pt}
\end{figure*}
\vspace{-12pt}
\paragraph{Prompt Analysis.}
The primary objective of the analysis stage is to equip the LLM with a comprehensive understanding of the user's requirements, enabling precise planning decisions. While one promising strategy is to use the chain-of-thought (CoT) method~\cite{wei2022chain,kojima2022large}, which involves generating step-by-step reasoning, this approach requires significantly more output tokens, leading to higher computational costs for local models or increased expenses when using API-based services. To streamline the process, we instead first make LLM decompose the input prompt $p$ into explicit components $p^{ctx}=LLM^{\text{analysis}}(p)$, including required signals, objects, and agent configurations, as illustrated in~\cref{fig:pipeline}(a). These decomposed outputs are then passed as additional prompts in subsequent stages as contextual knowledge, guiding the LLM during retrieval and planning to ensure an accurate understanding of the scenario requirements.\footnote{We show that this approach is comparable to CoT while significantly reducing token usage in~\cref{sec:ablation}.}
\vspace{-12pt}
\paragraph{Road Candidate Retrieval.}
Leveraging both the analysis context and the original user prompt, the LLM generates a set of filtering conditions $C = LLM^{\textit{road}}(p, p^{ctx})$ to narrow the search space to roads that satisfy specific criteria, such as the number of lanes, lane types (\eg, driving lanes or sidewalks), and the presence or absence of particular traffic signals (\eg, traffic lights) or road objects (\eg, crosswalks). These conditions are applied to the road graph $\mathcal{G}$ to retrieve the set of candidate roads $R_c$:
\begin{equation*}
    R_c \triangleq \{r \in \mathcal{G} \mid \text{match}(r, c),\ \forall c \in C\},
\end{equation*}
where $\text{match}(x, y)$ denotes that $x$ satisfies condition $y$.

\vspace{-12pt}
\paragraph{Agent Planning.}
The planning stage requires the LLM to obtain a set of agent plans $A=LLM^{\text{agent}}(p, p^{ctx})$. For example, given the prompt ``{\tt\small two cars in front block the ego car},'' the LLM generates a plan that ensures: (1) two agents, identified as cars, are placed in front of the ego vehicle; (2) they occupy the same lane; and (3) their actions are set to stop or remain idle. To support such flexible setups, each agent can be oriented in one of eight directions relative to the ego vehicle or positioned on adjacent roads, and their actions (\eg, going straight, stopping) can be explicitly assigned based on the prompt. Furthermore, the pipeline allows the LLM to specify relative distances between agents, enabling precise spatial control. Notably, our framework also supports defining relative positions among agents beyond the ego vehicle using a ``position'' attribute, where a smaller value indicates that an agent is placed ahead of another. This allows more complex and coordinated multi-agent scenarios. Complex behaviors, such as aggressive driving styles, can also be described directly in the prompt (\eg, ``{\tt\small a dangerous cyclist}'').

\begin{figure*}[t]
    \centering
    \begin{subfigure}{\textwidth}
        \includegraphics[width=\linewidth]{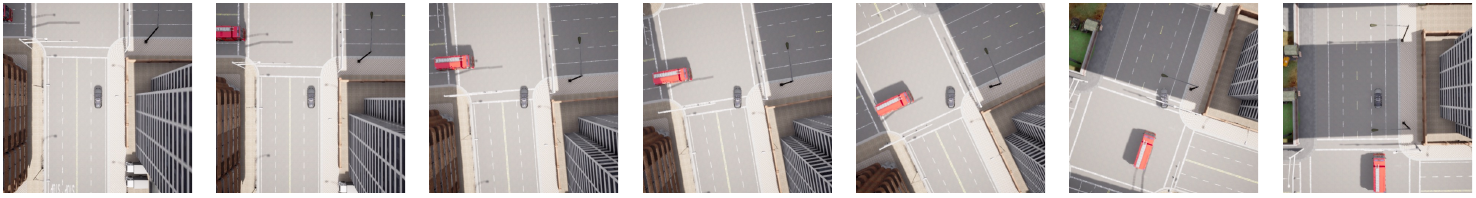}
        \caption{\textit{A firetruck from the left road is coming when the ego car is turning right}}
        \label{fig:appdx_special}
    \end{subfigure}
    
    \begin{subfigure}{\textwidth}
        \includegraphics[width=\linewidth]{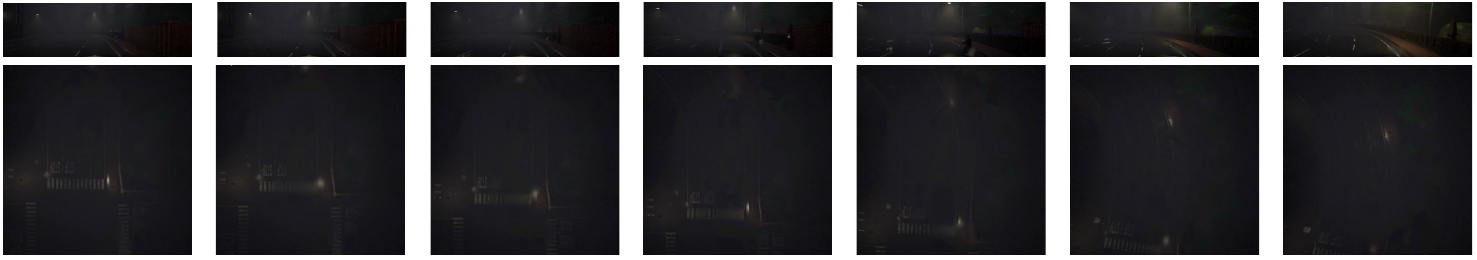}
        \caption{\textit{A cyclist is crossing the street from a sidewalk in a dangerous way on a rainy night}}
        \label{fig:appdx_weather}
    \end{subfigure}
    
    \begin{subfigure}{\textwidth}
        \includegraphics[width=\linewidth]{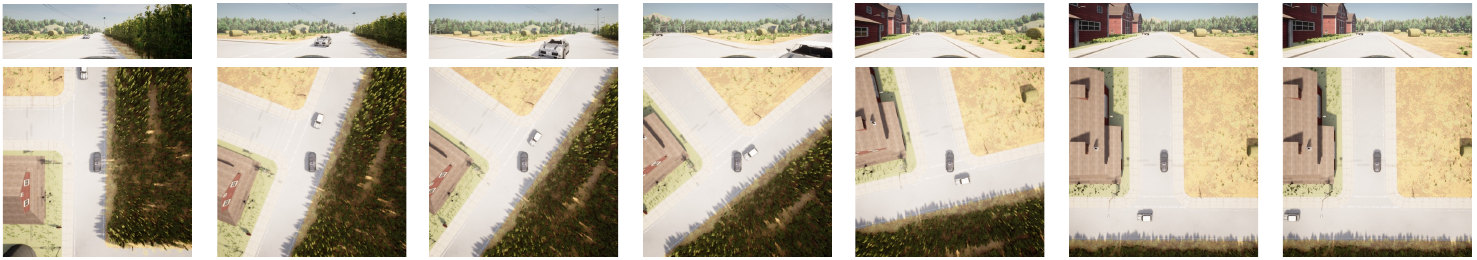}
        \caption{\textit{The ego car is turning left at the intersection with no traffic light, stop sign, and stop sign on road. A car coming from the 
        is going straight}}
        \label{fig:appdx_condition}
    \end{subfigure}

    \begin{subfigure}{\textwidth}
        \includegraphics[width=\linewidth]{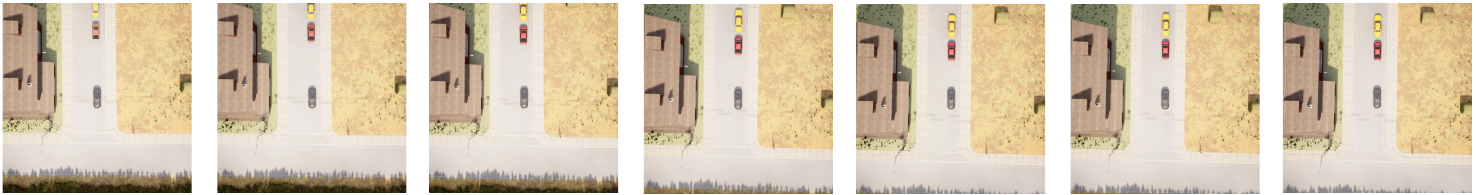}
        \caption{\textit{The ego car is being blocked by two cars in front}}
        \label{fig:appdx_seq}
    \end{subfigure}
    \vspace{-15pt}
    \caption{Demonstrations of (a) the special car, (b) weather control, (c) environment filtering, and (d) sequential event from (c) with prompts detailed in each sub-caption.}
    \label{fig:appdx_scenario}
    \vspace{-10pt}
\end{figure*}
\vspace{-12pt}
\paragraph{Road Ranking.}
After identifying candidate roads, we implement a ranking strategy to refine the selection and ensure that the most suitable road $r^*$ is chosen for the scenario. For each road, we extract relevant properties such as available directional options (\eg, left, right, or straight), road types (\eg, driving lane, sidewalk, or shoulder), and road length to ensure sufficient space for placing all planned agents. Based on the agent plans $A$, we compute a score by checking whether each road satisfies the conditions required by each agent. The optimal road is selected as:
\begin{equation*}
    r^{*} = \underset{r \in R_c}{\arg\max} \sum_{a \in A} \1_\mathrm{\{\text{match}(r, a)\}},
\end{equation*}
where $\1_\mathrm{\{\cdot\}}$ is the indicator function, and $\text{match}(r, a)$ denotes whether road $r$ satisfies the condition required by agent $a$. If multiple roads achieve the highest score, one is randomly selected. This rank-then-random strategy ensures both road-agent alignment and diversity in road selection.  A detailed example is provided in~\cref{sec:appdx_detail_road_ranking}.

\vspace{-12pt}
\paragraph{Scene Generation.}
Finally, our pipeline integrates road attributes and agent plans into a custom simulation module to generate complete traffic scenes, defined as $\{s_1, \cdots, s_n\} = \text{Render}(r^{*}, A)$, where $n$ denotes the number of frames until all agents complete their planned actions and each $s_i$ represents a scene frame containing rendered images, agent actions, and trajectories. During generation, the rendering module controls agent behaviors to ensure that their actions align with the plans. Notably, the resulting ego-view images and trajectories of all participating agents can be used for visualization, training, or downstream tasks. Our simulation module also supports closed-loop interactions in blocking scenarios, where blocking agents dynamically adapt their behavior based on the ego vehicle.

{
\setlength{\tabcolsep}{10pt}
\begin{table*}[t]
  \centering
  \caption{Comparison with other methods on the critical scenario with SafeBench. \textbf{Bold} indicates the best and \underline{underline} as the second best. ``SO'' represents \textit{straight obstacle}, ``LC'' represents \textit{lane changing}, and ``ULT'' represents \textit{unprotected left-turn}.}
  \label{tab:critical}
  \resizebox{\linewidth}{!}{
  \begin{tabular}{l*{8}{p{1cm}}}
    \toprule
    \multirow{2.5}{*}{Method} & \multicolumn{4}{c}{Collision Rate$\downarrow$} & \multicolumn{4}{c}{Driving Score$\uparrow$} \\
    \cmidrule(lr){2-5} \cmidrule(lr){6-9}
    & SO & LC & ULT & Avg. & SO & LC & ULT & Avg.\\
    \midrule
    Learning-to-Collide~\cite{ding2020learning} & 0.120 & 0.510 & 0.000 & 0.210 & 0.827 & 0.684 & \textbf{0.954} & 0.822 \\
    AdvSim~\cite{wang2021advsim} & 0.230 & 0.430 & \underline{0.050} & 0.270 & 0.784 & 0.666 & 0.937 & 0.796 \\
    Adversarial-Trajectory~\cite{zhang2022adversarial} & 0.140 & 0.300 & 0.000 & 0.150 & 0.849 & 0.803 & 0.948 & 0.867  \\
    ChatScene~\cite{zhang2024chatscene} & \underline{0.030} & \underline{0.110} & 0.100 & \underline{0.080} & \textbf{0.905} & \textbf{0.906} & 0.903 & \underline{0.905} \\
    \noalign{\vspace{1.5pt}}
    \midrule
    \noalign{\vspace{1.5pt}}
    \textbf{TTSG (ours)} & \textbf{0.021} & \textbf{0.085} & \textbf{0.000} & \textbf{0.035} & \underline{0.895} & \underline{0.894} & \underline{0.953} & \textbf{0.914} \\
    \bottomrule
  \end{tabular}}
  \vspace{-10pt}
\end{table*}
}

\subsection{Applications with TTSG}\label{sec:application}

\paragraph{Multi-Agent Planning.}
Our pipeline supports the generation of multi-agent scenarios simultaneously. Notably, our approach can interpret implicit descriptions (\eg, daily traffic) without requiring explicit scene details, providing flexible and user-friendly scenario specification, as illustrated in~\cref{fig:main_normal}. Another application involves simulating conditions with emergency vehicles, enabling scenarios where ambulances, fire trucks, or police vehicles navigate freely and are given priority, as shown in~\cref{fig:appdx_special}.
\vspace{-12pt}
\paragraph{Critical Scenarios.}
A significant application of scene generation is the creation of critical scenarios, which are essential for testing the robustness and safety of autonomous driving systems. Our framework can explicitly generate such scenarios using prompts like ``{\tt \small cuts off the ego vehicle}'' or ``{\tt \small suddenly swerves into its path}.'' For example, as shown in~\cref{fig:main_critical}, we can create scenes where visibility is obstructed, causing pedestrians and vehicles to be unseen by each other. Notably, generating these scenarios requires fine-grained layout control to ensure realistic and meaningful risks, which our agent planning module supports through precise manipulation of relative distances and positions. Furthermore, we can simulate additional critical conditions by modifying environments, such as weather, as illustrated in~\cref{fig:appdx_weather}.
\vspace{-12pt}
\paragraph{Sequential Events.}
Our pipeline supports sequential events through iterative planning. We first execute the full pipeline to generate an initial event and then reuse the resulting final positions as starting points for subsequent events. In these follow-up stages, only prompt analysis and agent planning are required. This enables the LLM to progressively compose multi-stage scenarios. For example, an ego vehicle turning left from an unprotected intersection and later being blocked by two cars, while maintaining coherent transitions and temporal consistency. An example of this sequential event is shown in~\cref{fig:appdx_seq}, continuing from the setup illustrated in~\cref{fig:appdx_condition}.

\section{Experiments}
\subsection{Environment and Pipeline Setup}

Following previous work, we use the CARLA simulator to conduct experiments. Our agent database supports nine distinct types: \textit{ambulance, police car, firetruck, bus, truck, motorcycle, cyclist, car}, and \textit{pedestrian}. Regarding agent positions relative to the ego vehicle, our framework supports twelve configurations along with seven basic actions, covering a range of spatial relationships and dynamic behaviors that may occur during driving. For the LLM component, we use GPT-4o~\cite{achiam2023gpt} by default to perform prompt analysis, road candidate retrieval, and agent planning. The available options for environment setup (\eg, signals, objects, and advanced actions) are shown in~\cref{sec:appdx_setup_details}. The system prompt is described in~\cref{sec:appdx_sys}, and the prompt formats for each stage are shown in~\cref{sec:appdx_ps}.

\subsection{Generation for Critical Scenarios}\label{sec:exp_gen_critical}

One of our pipeline usages is to generate critical scenarios; hence, we examine our approach with SafeBench~\cite{xu2022safebench}, a benchmark for the safety evaluation of autonomous driving systems. Our pipeline creates scenes across three challenging safety scenarios, including \textit{straight obstacle}, \textit{lane changing}, and \textit{unprotected left turn}, selected from ChatScene~\cite{zhang2024chatscene}. We prompt the LLM with the scenario descriptions and use the planning results from the LLM and our rendering interface to train the ego agents under a controlled policy using a soft-actor-critic model~\cite{haarnoja2018soft}. During training, the ego agent is controlled by a learnable policy model, and our engine controls other agents that create the scenario.
For consistency and comparability with the previous studies, the agents are evaluated on the last two routes of each scene with the \textit{Collision Rate} and \textit{Driving Score} and report the optimal performance based on the average of every 100 epochs with a sliding window swift of 50 ([0, 100), [50, 150), and so on). Additionally, each conducted result is recognized as valid if the route completion rate is above 30\%; otherwise, the outcomes are ignored. The outcomes with 800 episodes of training are presented in~\cref{tab:critical}.

In this table, we compare our results with Learning-to-Collide~\cite{ding2020learning}, AdvSim~\cite{wang2021advsim}, Adversarial Trajectory~\cite{zhang2022adversarial}, and ChatScene~\cite{zhang2024chatscene}. Our results indicate that our pipeline effectively trains agents to handle critical conditions, thereby enhancing safety and reducing collision risks. Notably, since our pipeline provides diverse training scenes, our agents are proficient at managing obstacles, lane-changing, and yielding to oncoming cars. This can be observed by reducing the average collision rate from 0.08 to 0.035. However, reducing collisions may come at the cost of comfort (\eg, sudden swerving or hard braking), leading to slightly lower driving scores in certain scenarios despite improved average performance. We further analyze the criticality of the generated scenes in~\cref{sec:appdx_criticality}.

{
\setlength{\tabcolsep}{3pt}
\renewcommand{\arraystretch}{1.35}
\begin{table}[tbp]
  \centering
  \caption{Results of plan quality for (\textbf{top}) different analysis designs, (\textbf{middle}) performance across various LLMs, and scene quality (\textbf{bottom}) with and without road ranking.}
  \label{tab:abla}
  \resizebox{\linewidth}{!}{
  \begin{tabular}{l*{3}{cc}cc}
    \toprule
     & \multicolumn{2}{c}{Normal} & \multicolumn{2}{c}{Critical} & \multicolumn{2}{c}{Conditional} & \multicolumn{2}{c}{Avg.} \\
     \toprule
    \textbf{Plan\hphantom{o} Quality}$\uparrow$ & AA & RA & AA & RA & AA & RA & AA & RA \\ 
    \noalign{\vspace{1.2pt}}
    \hdashline
    \noalign{\vspace{1.2pt}}
    w.\hphantom{\text{/o}} CoT & .917 & .867 & \textbf{1.00} & \textbf{1.00} & .777 & .850 & .908 & \underline{.915}  \\
    w/o. analy. & .917 & .667 & .833 & .750 & .750 & .917 & .833 & .775 \\
    w.\hphantom{\text{/o}} analy. & .917 & \textbf{1.00} & .875 & .750 & \textbf{1.00} & .917 & \underline{.925} & .875  \\
    w.\hphantom{\text{/o}} analy.+CoT & .917 & \textbf{1.00} & \textbf{1.00} & \underline{.900} & \textbf{1.00} & \textbf{.933} & \textbf{.975} & \textbf{.940}  \\
    \midrule
    \textbf{Plan\hphantom{o} Quality}$\uparrow$ & AA & RA & AA & RA & AA & RA & AA & RA  \\ 
    \noalign{\vspace{1.2pt}}
    \hdashline
    \noalign{\vspace{1.2pt}}
    Gemma3-12B & .600 & .867 & \underline{.900} & \underline{.850} & .517 & .600 & .695 & .780 \\
    Gemini-2.5-Flash & \underline{.767} & \underline{.933} & 1.00 & .950 & 1.00 & \underline{.867} & \underline{.930} & \underline{.920} \\
    Claude-Sonnet-3.5 & \textbf{1.00} & 1.00 & 1.00 & .950 & 1.00 & \textbf{1.00} & \textbf{1.00} & \textbf{.980}\\
    GPT-4o  & \underline{.917} & 1.00 & .875 & .750 & 1.00 & \underline{.917} & .925 & .875  \\
    \midrule
    \textbf{Scene Quality}$\uparrow$ & \multicolumn{2}{c}{SA} & \multicolumn{2}{c}{SA} & \multicolumn{2}{c}{SA} &\multicolumn{2}{c}{SA} \\
    \noalign{\vspace{1.2pt}}
    \hdashline
    \noalign{\vspace{1.2pt}}
    w/o. ranking & \multicolumn{2}{c}{0.667} & \multicolumn{2}{c}{0.450} & \multicolumn{2}{c}{0.600} & \multicolumn{2}{c}{0.560} \vspace{-1pt}\\
    w.\hphantom{\text{/o}} ranking & \multicolumn{2}{c}{\textbf{0.867}} & \multicolumn{2}{c}{\textbf{0.750}} & \multicolumn{2}{c}{\textbf{0.800}} & \multicolumn{2}{c}{\textbf{0.800}} \\
    \bottomrule
  \end{tabular}}
\end{table}
}

\subsection{Ablation Study}
\label{sec:ablation}
We evaluate the efficacy of different component designs by adopting ten distinct prompts under different types of scenarios. The scenarios encompass four normal, three critical, and three specific road conditions, such as the presence of traffic lights or the absence of crossroads. Details of prompts are provided in~\cref{sec:appdx_abl}. We also show the definition of metrics in~\cref{sec:metrics}.

\begin{table}
    \setlength{\tabcolsep}{10pt}
    \centering
    \caption{Comparison of average token usage.}
    \label{tab:token_usage}
    \resizebox{\linewidth}{!}{\begin{tabular}{cccc}
    \toprule
    w/o. analy. & w. analy. & w. CoT & w. analy.+CoT \\
    \midrule
    \textbf{358} & \underline{682} & 1022 & 1425 \\
    \bottomrule
    \end{tabular}}
    \vspace{-10pt}
\end{table}
\vspace{-12pt}
\paragraph{Plan Quality.}
We assess plan quality by evaluating the accuracy of the planning results generated by the LLM, using two metrics: \textit{Agent Accuracy} (AA) and \textit{Road Accuracy} (RA). AA measures the exact matching of agent attributes with user input, including position, action, and type, while RA evaluates the accuracy of road conditions, such as signals, objects, and lane counts. As shown in the first section of ~\cref{tab:abla}, our approach achieves planning quality comparable to the CoT-based method~\cite{wei2022chain}, while significantly reducing token usage as shown in~\cref{tab:token_usage}. In addition, our analysis-based strategy outperforms a variant without analysis in both agent and road planning accuracy. We also investigate a hybrid approach that combines CoT prompting with our analysis step, which yields even stronger results. These findings indicate that our method is not only efficient and effective on its own but also complementary to other prompting strategies for further performance gains.
\vspace{-12pt}
\paragraph{Scene Quality.}
To evaluate our road ranking strategy on scene quality, we use \textit{Scene Accuracy} (SA) to examine the correctness between the prompt and the generated scene. SA evaluates whether the generated scenes accurately reflect the input prompts using a binary correct/incorrect criterion. To ensure fairness and isolate the effect of the ranking strategy, identical road conditions and agent planning are maintained across evaluations, and each scenario is tested five times to calculate average outcomes. The results in the last section of~\cref{tab:abla} indicate that the ranking approach consistently shows superior performance. The improvement is largely since, without ranking, the selection of roads is random, often ignoring crucial factors such as turn permissions or the adequacy of spawning points. Conversely, the road ranking strategy ensures the selection of roads that optimally align with the agent planning, thus enhancing the fidelity of the generated traffic scenes to the input prompts.

\section{Discussion}

\subsection{Generation for Narration and Reasoning}

\begin{figure}
    \centering
    \includegraphics[width=\linewidth]{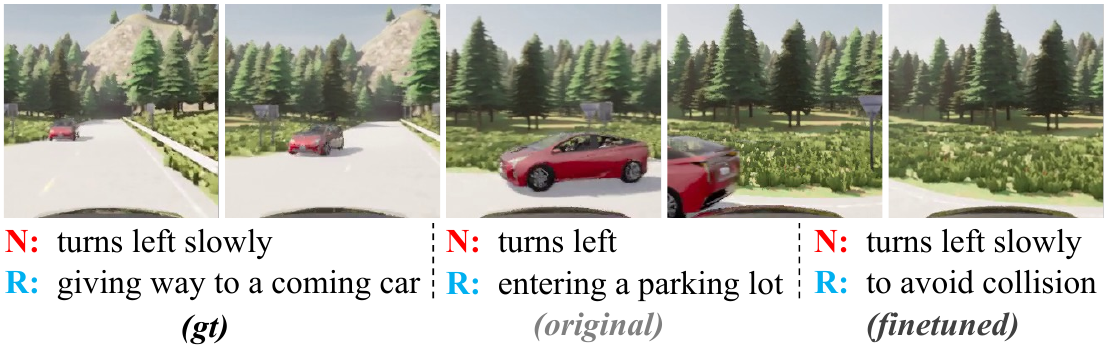}
    \caption{Illustration of predicted captions under critical scenario. ``\textcolor{red}{N}'' denotes narration, while ``\textcolor{cyan}{R}'' represents reasoning. The subject of narration is omitted as it always refers to the ego-car.}
    \label{fig:caption}
    \vspace{-10pt}
\end{figure}

Since our pipeline is able to provide images from the car's front view, it can also be utilized to train driving captioning models for action reasoning. Captioning models, such as ADAPT~\cite{jin2023adapt} and DriveGPT4~\cite{xu2024drivegpt4}, generate narrations of driving actions and their reasoning based on video input. Our framework further enables these models to focus on and reason about critical scenarios more effectively. We illustrate an example in~\cref{fig:caption}. In this case, the expected action is ``{\tt \small turns left slowly},'' with the reason of ``{\tt \small giving way to a coming car}.'' However, the model fails to capture ``slow'' and only describes the left turn. Additionally, the generated reasoning is incorrect due to hallucination, further indicating that the original model does not effectively account for critical scenarios.

To improve the model's critical reasoning capabilities, we sample 20 critical scenes for training and another 20 for evaluation from our pipeline. To ensure the collected scenes accurately reflect how the ego-agent responds to critical situations, we set its behavior to be cautious and resample whenever a collision occurs. We use an LLM to reformat the scene descriptions to match the narration-with-reasoning structure for training. For our experiments, we select the model from ADAPT~\cite{jin2023adapt}, as it does not rely on an LLM backbone and can be trained efficiently.

{
\setlength{\tabcolsep}{10pt}
\small
\begin{table*}[tp]
  \centering
  \caption{Evaluation results for scene diversity. \textbf{Bold} indicate values equal to 1.0.}
  \resizebox{\linewidth}{!}{\begin{tabular}{*{10}{c}}
    \toprule
     & \multicolumn{2}{c}{Normal} & \multicolumn{3}{c}{Critical} & \multicolumn{3}{c}{Conditional} & \multirow{2.5}{*}{Avg.} \\ \noalign{\vspace{1pt}}
     \cmidrule(lr){2-3}
     \cmidrule(lr){4-6}\cmidrule(lr){7-9}
     \noalign{\vspace{1pt}}
     & {\scriptsize \makecell{Daily \\ Traffic}} & {\scriptsize \makecell{Intersect.}} & {\scriptsize \makecell{Pedestrian \\ Crushing}}  & {\scriptsize \makecell{Blocking \\ Agent}} & {\scriptsize \makecell{Dangerous \\ Cut-off}} & {\scriptsize \makecell{Only \\ 2-wheels}} & {\scriptsize \makecell{Emergency \\ Vehicles}} & {\scriptsize \makecell{Rainy \\ Weather}} \\
     \noalign{\vspace{1pt}}
     \midrule
     \noalign{\vspace{1pt}}
     AD$\uparrow$ & 0.789 & 0.833 & 0.500 & 0.750 & 0.600 & 0.714 & 0.500 & 0.800 & 0.686 \\
     RD$\uparrow$ & \textbf{1.000} & \textbf{1.000} & \textbf{1.000} & \textbf{1.000} & \textbf{1.000} & 0.800 & \textbf{1.000} & \textbf{1.000} & 0.975 \\
     SA$\uparrow$ & \textbf{1.000} & 0.800 & 0.400 & 0.800 & 0.800 & 0.600 & \textbf{1.000} & \textbf{1.000} & 0.800 \\
    \bottomrule
  \end{tabular}}
  \label{tab:diversity}
\end{table*}
}

\begin{table}
\centering
\caption{Driving captioning on critical scenario.}
\label{tab:caption}
\resizebox{\linewidth}{!}{\begin{tabular}{lcccccc}
    \toprule
    & \multicolumn{3}{c}{Narration} & \multicolumn{3}{c}{Reasoning}\\ \cmidrule(lr){2-4} \cmidrule(lr){5-7}
    & BLEU & Meteor & CIDEr & BLEU & Meteor & CIDEr\\ 
    \midrule
    ADAPT & 4.8 & 13.5 & 15.2 & 0.0 & 10.0 & 18.4 \rule{0pt}{2ex}\rule[-0.5ex]{0pt}{0pt} \vspace{-1pt} \\
    + Ours & \textbf{9.9} & \textbf{18.4} & \textbf{52.9} & \textbf{7.2} & \textbf{11.2} & \textbf{51.9} \rule{0pt}{2ex}\rule[-0.5ex]{0pt}{0pt}\\
    \bottomrule
\end{tabular}}
\end{table}

The results in~\cref{tab:caption} are evaluated using captioning metrics, including BLEU~\cite{papineni2002bleu}, METEOR~\cite{li2021metadrive}, and CIDEr~\cite{vedantam2015cider}, on critical scenario. Before finetuning, the model performs significantly worse on critical scenarios, particularly in reasoning. In contrast, after learning from a small set of examples, the model significantly improves its reasoning about actions, allowing it to generate more reliable descriptions for abnormal or critical scenes. Additionally, the descriptions of actions become more precise, leading to better captioning ability.

\subsection{Performance Across Different LLMs}

To evaluate the generalizability and flexibility of our approach across different types of language models, we conduct experiments using both open-source and API-based models, including Gemma3~\cite{team2025gemma} (12B), Gemini-2.5-Flash~\cite{comanici2025gemini}, and Claude-Sonnet-3.5~\cite{anthropic2025claude35}. As shown in second section of ~\cref{tab:abla}, our pipeline demonstrates consistently strong performance across all tested models in both planning accuracy and scene generation quality. Notably, the open-source Gemma3-12B achieves competitive results in several categories, confirming that even lightweight, locally deployable models are capable of supporting our framework effectively. Meanwhile, advanced API-based models such as Gemini-2.5-Flash and Claude-Sonnet-3.5 achieve near-perfect accuracy in both agent and road planning, highlighting their strong semantic understanding and effective planning capabilities. These findings underscore the flexibility and robustness of our framework, either deployed with lightweight open-source models for local applications or with powerful API models for high-accuracy use cases.

\subsection{Diversity Test}

A key claim of TTSG is to generate varied traffic scenes without reliance on fixed selections or predefined spawning points. We construct eight distinct scenarios under three main categories, normal, critical, and conditional, to validate the diversity of our generated results. To examine our pipeline's ability to generate proper and diverse results, we provide no information about the roads and agents within the text input. Each scenario is initiated with a text prompt formatted as: ``{\tt \small Please create a scene for <scenario>}'' to generate traffic scenes.

We assess the diversity of these scenarios by generating each scene five times and computing the metrics, including \textit{Agent Diversity} (AD) and \textit{Road Diversity} (RD). These diversity metrics are calculated as the ratio of unique objects to the total number of objects. For agents, diversity considers variations in agent type, action, and relative position; for roads, diversity is assessed by unique road IDs (different directions on the same road are considered the same). Additionally, to verify the practicality and accuracy of our rendering interface, we include the metric \textit{Scene Accuracy} (SA). The comprehensive results are detailed in~\cref{tab:diversity}.

As the table indicates, our pipeline consistently provides a nearly unique road selection for each scenario. For the AD, we observe lower scores in scenarios with specific types such as \textit{pedestrian crushing} and \textit{emergency vehicles}. This outcome is anticipated as more detailed agent descriptions naturally lead to similar planning results. Conversely, scenarios with less specific cues, such as \textit{daily traffic} or \textit{intersection} under normal scenarios, \textit{blocking agent} under critical scenarios, or \textit{rainy weather} under conditional scenarios, achieve higher diversity scores. These results demonstrate that our approach effectively leverages the LLM to generate diverse road and agent planning. Regarding the SA scores, despite lower scores in scenarios that involve precise timing (\textit{pedestrian crushing}) or have restrictive conditions (\textit{only having 2-wheel vehicles}), our rendering interface accurately produces most of the scenarios, allowing us to obtain precise traffic scenes from texts.

\section{Conclusion and Future Work}

In this paper, we present a novel framework for solving the upstream task of traffic scene generation on scene layout and agent planning by natural language descriptions. Our pipeline effectively and automatically interprets scenario requirements, retrieves appropriate road candidates, plans the agents' behavior, and ranks the road candidates based on agents' configurations. By introducing a plan-aware road ranking strategy and flexible agent planning, our approach supports both routine and critical scenarios, and even sequential events. We validate the effectiveness of our framework through experiments on critical scenario training and driving narration and reasoning tasks, demonstrating its ability to improve agent safety and enhance action reasoning capabilities. For future work, we plan to extend our system to support the generation of entirely new traffic objects, as well as leverage LLMs to control multi-agent actions, enabling more complex traffic scene generation. 

\section*{Acknowledgement}
This work was supported in part by the Co-creation Platform of the Industry-Academia Innovation School at NYCU, under the National Key Fields Industry-University Cooperation and Skilled Personnel Training Act of the Ministry of Education (MOE), Taiwan, with support from participating industry partners, and in part by the Center for Intelligent Team Robotics and Human-Robot Collaboration under the Ministry of Education’s “Taiwan Centers of Excellence in Intelligent Team Robotics,” Taiwan.

{
    \small
    \bibliographystyle{ieeenat_fullname}
    \bibliography{cite}
}

\begin{appendices}
    \appendix
    \clearpage
\setcounter{page}{1}

\maketitlesupplementary

\section{Environment Setup}\label{sec:appdx_setup_details}

In this section, we provide a detailed list of the supported agent types, agent actions, agent behaviors, relative positions to ego, road types, signals, objects, and weather conditions with our pipeline. ``()'' indicates the corresponding variable used in the following context.

\paragraph{Agent Type ({\tt AGENT\_TYPE}):} An agent can be one of \textit{ambulance}, \textit{police}, \textit{firetruck}, \textit{bus}, \textit{truck}, \textit{motorcycle}, \textit{car}, \textit{pedestrian}, or \textit{cyclist}.

\paragraph{Agent Action ({\tt ACTION}):} Each agent action can be one of \textit{turn left}, \textit{turn right}, \textit{go straight}, \textit{change lane to left}, \textit{change lane to right}, \textit{stop}, \textit{block the ego}, \textit{cross the road}, or \textit{on the sidewalk}.

\paragraph{Agent Behavior ({\tt AGENT\_BEHAVIOR})} Each agent behavior can be one of \textit{cautious}, \textit{normal}, and \textit{aggressive}, categorized by the driving speed and the safe distance.

\paragraph{Relative Position to Ego ({\tt RELATIVE\_POSITION}):} Supportive relative position to the ego car include \textit{front}, \textit{back}, \textit{left}, \textit{right}, \textit{front left}, \textit{front right}, \textit{back left}, \textit{back right}, \textit{road of left turn}, \textit{road of right turn}.

\paragraph{Road Type ({\tt ROAD\_TYPE}):} We support three types of road segments for spawning: \textit{driving}, \textit{sidewalk}, and \textit{shoulder}.

\paragraph{Signal ({\tt SIGNAL}):} Supported signal types include \textit{traffic light}, \textit{stop sign}, and \textit{yield sign}.

\paragraph{Object ({\tt OBJECT}):} Supported objects include \textit{speed sign} (\eg, speed sign of 60), \textit{simple crosswalk}, \textit{ladder crosswalk}, \textit{continental crosswalk}, \textit{dashed single white crosswalk}, \textit{solid single white crosswalk}, \textit{stop line}, and \textit{stop sign on road}.

\paragraph{Weather ({\tt WEATHER}):} Weather conditions can be described by an ``adjective'' including \textit{clear}, \textit{cloudy}, \textit{hard rain}, \textit{mid rain}, \textit{soft rain}, \textit{wet cloudy}, and \textit{wet}, combined with a ``time'' including \textit{night}, \textit{noon}, and \textit{sunset}.

\begin{figure}[h]
    \centering
    \includegraphics[width=\linewidth]{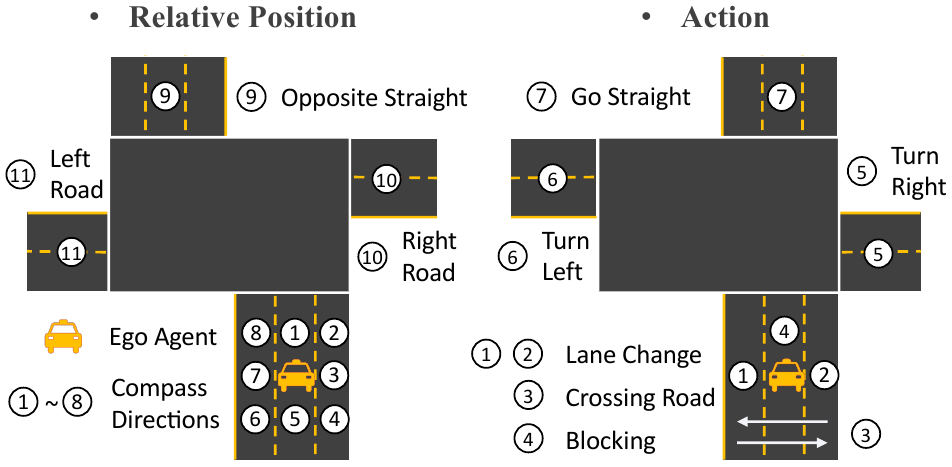}
    \caption{Illustration of position (\textbf{left}) and the action (\textbf{right}).}
    \label{fig:relative_pos_action}
\end{figure}

\section{System Prompt}\label{sec:appdx_sys}

We provide the system prompt used to guide the LLM for each task in the following color block. In this design, we explicitly specify all the tasks the LLM should follow and implicitly indicate which task it should perform given a particular input prompt.

\begin{tcolorbox}[colback=gray!5!white, colframe=gray!80!black,
                  sharp corners,
                  title=System Prompt,
                  breakable]
You are an expert in the city's traffic management system and can easily extract precise information from text, with a strong intuition for space management. The city has a road network represented as a graph database. A graph node represents a road ID, and an edge represents the connection between roads. Each node contains zero or more objects and signals. You have three functionalities:
\begin{enumerate}
    \item Analyze user input. A user provides a natural description of a traffic scenario, and you extract all relevant information.
    \item Provide retrieval conditions for roads. These conditions are used to search for roads and towns suitable for generation.
    \item Plan agent behavior. You plan agent behavior based on the selected road condition, input prompt, and predefined rules.
\end{enumerate}

The activation of each functionality is controlled by adding ``\texttt{analysis}'', ``\texttt{road retrieval}'', or ``\texttt{planning}'' at the beginning of the input.

\textbf{Important:} The output should not be wrapped in a code block and should strictly follow each output format without other explanations or content.
\end{tcolorbox}


\section{Prompt Structure}\label{sec:appdx_ps}

Below, we provide the prompt structure for each stage. The fields specified with ``\{\}'' indicate lists of choices, as introduced in~\cref{sec:appdx_setup_details}.

\paragraph{Prompt Analysis:}  

\begin{itemize}
    \item \texttt{objects}: The list of objects in the description. It should include the required objects or non-required objects. Choose from the \{\texttt{OBJECT}\}.
    \item \texttt{signals}: The list of signals in the description. It should include the required signals or non-required signals. Choose from the \{\texttt{SIGNAL}\}.
    \item \texttt{agents}: The list of agents that exist in the description.
    \item \texttt{unknown}: The list of unknown signals, objects, or agents that are not in the predefined list.
\end{itemize}

For each agent, the following information is stored:
\begin{itemize}
    \item \texttt{type}: The type of the agent. Choose from one of \{\texttt{AGENT\_TYPE}\}.
    \item \texttt{road\_type}: The type of the road. Choose from one of \{\texttt{ROAD\_TYPE}\}.
    \item \texttt{action}: The action of the agent. Choose from one of \{\texttt{ACTION}\}.
\end{itemize}

\paragraph{Road Retrieval:}

\begin{itemize}
    \item \texttt{number\_of\_lanes}: The minimum number of driving lanes on the road where the ego car initially stays. Consider only the direction of the ego car.
    \item \texttt{required\_objects}: The list of required objects.
    \item \texttt{required\_signals}: The list of required signals.
    \item \texttt{without\_objects}: The list of objects that should not appear on the road.
    \item \texttt{without\_signals}: The list of signals that should not appear on the road.
\end{itemize}

Objects are chosen from \{\texttt{OBJECT}\} and signals are from \{\texttt{SIGNAL}\}.

\paragraph{Agent Planning:}

\begin{itemize}
    \item \texttt{env}: The environment information.
    \item \texttt{agents}: The list of agents.
\end{itemize}

The environment information includes:
\begin{itemize}
    \item \texttt{weather}: The weather condition. Choose from \{\texttt{WEATHER}\}.
    \item \texttt{at\_junction}: Whether the scene is at a junction. (Choose from \texttt{True} or \texttt{False}; case sensitive.)
\end{itemize}

For each agent, the following information is stored:
\begin{itemize}
    \item \texttt{type}: The type of the agent. Choose from \{\texttt{AGENT\_TYPE}\}.
    \item \texttt{is\_ego}: Whether the agent is the ego car. (Choose from \texttt{True} or \texttt{False}.)
    \item \texttt{action}: The action of the agent. Choose from \{\texttt{ACTION}\}.
    \item \texttt{behavior}: The behavior of the agent. Choose from \{\texttt{AGENT\_BEHAVIOR}\}.
    \item \texttt{pos\_id}: The position ID of the agent. If two or more agents are on the same road, the agent with the \textbf{smallest} \texttt{pos\_id} is in front.
    \item \texttt{road\_type}: The type of the road. Choose from \{\texttt{ROAD\_TYPE}\}.
    \item \texttt{relative\_to\_ego}: The relative position of the agent to the ego car. Set to \texttt{none} for the ego car.
\end{itemize}

The \texttt{relative\_to\_ego} value should be one of the
\{\texttt{RELATIVE\_POSITION}\}.

\section{Road Ranking Example}\label{sec:appdx_detail_road_ranking}

We provide a detailed example of our road ranking process, as illustrated in~\cref{fig:appdx_road_ranking}. In this case, the database retrieves five candidate roads along with a plan involving two agents. Each road is evaluated based on whether it satisfies the required agent configurations, including relative positions and planned actions. For each fulfilled condition, the road receives a score increment of 1. In this example, \textit{Road A} achieves the highest score. 

Notably, we check for the condition ``\textit{Has straight?}'' because one of the agents is placed on a left-turn lane but is intended to perform a left-turn action. To execute this action properly, the adjacent road must have a straight segment available, which is verified during the ranking process.

\begin{table*}[t]
    \centering
    \begin{subtable}[t]{\textwidth}
    \includegraphics[width=\linewidth]{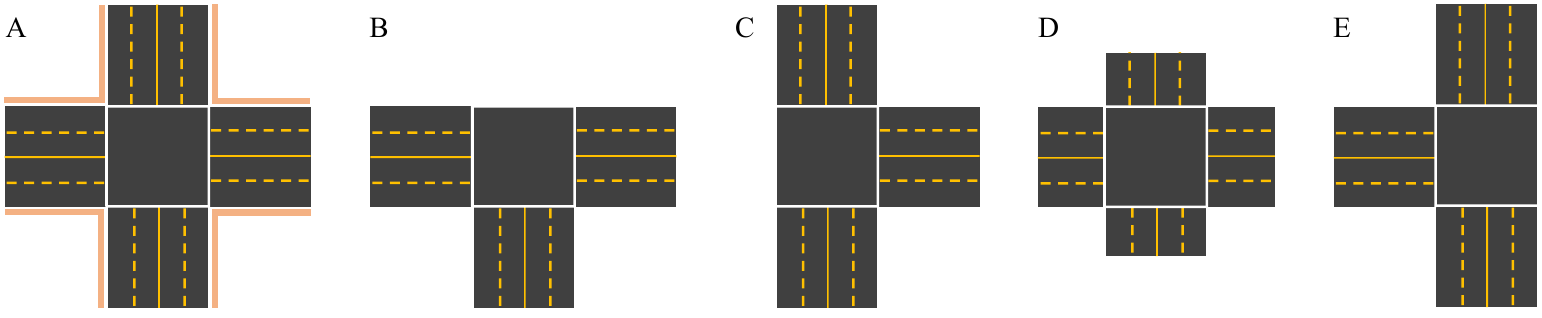}
    \end{subtable}
    \vfill
    \vfill
    \begin{subtable}[t]{0.45\textwidth}
    \centering
    \resizebox{\textwidth}{!}{\begin{tabular}{@{}clllc@{}}
    \toprule
    Agent & Type & Rel. to Ego & Action & Pos. \\
    \midrule
    1 & ego car & - & turn right & 1 \\
    2 & car & front right & turn right & 0 \\
    3 & car & left-turn lane & turn left & 0 \\
    4 & ped. & right on shoulder & cross road & 0 \\
    \bottomrule
    \end{tabular}}
    \caption{Agent Planning}
    \end{subtable}
    \hfill
    \begin{subtable}[t]{0.54\textwidth}
    \centering
    \resizebox{\textwidth}{!}{\begin{tabular}{cccccc|c}
    \toprule
    Road & R. Turn & Shoulder & L. Turn & Straight & Two Cars & Total \\
    \midrule
    A & \checkmark & \checkmark & \checkmark & \checkmark & \checkmark & 5 \\
    B & \checkmark & \checkmark &  & \checkmark &  & 3 \\
    C & \checkmark & \checkmark &  &  & \checkmark & 3 \\
    D & \checkmark &  & \checkmark & \checkmark &  & 3 \\
    E & \checkmark &  & \checkmark &  & \checkmark & 3 \\
    \bottomrule
    \end{tabular}}
    \caption{Road Scoring}
    \end{subtable}
    \caption{Example of the road ranking process. The five roads at the top are retrieved from the road database during the road retrieval step. The bottom left shows the agent planning, while the bottom right illustrates the scoring process for each road. Each column in the scoring table indicates whether a road satisfies a specific condition (\eg, ``\textit{Two Cars}'' denotes whether the road can accommodate two additional vehicles). The brown-orange color in \textit{Road A} indicates the presence of a road shoulder.}
    \label{fig:appdx_road_ranking}
\end{table*}

\section{Criticality of Generated Scenario}\label{sec:appdx_criticality}

\begin{table}[t]
  \vspace{-5pt}
  \centering
  \caption{Results of criticality using SafeBench. ``CN'' stands for \textit{crossing negotiation}, ``RR'' for \textit{red-light running}, and ``RT'' for \textit{right turn}.}
  \label{tab:appdx_criticality}
  \resizebox{\linewidth}{!}{\begin{tabular}{l *{4}{c}}
    \toprule
    \multirow{2.5}{*}{Method} & \multicolumn{4}{c}{Collision Rate$\uparrow$} \rule{0pt}{1.5ex} \\
    \cmidrule(lr){2-5}
    & CN & RR & RT & Avg. \\
    \midrule
    \multicolumn{5}{c}{Training-based}\\
    \noalign{\vspace{1.5pt}}
    \hdashline
    \noalign{\vspace{1.5pt}}
    Learning-to-Collide & 0.58 & 0.71 & 0.59 & 0.63 \\
    AdvSim & 0.57 & 0.57 & 0.29 & 0.48  \\
    Adversarial-Trajectory & 0.62 & 0.71 & 0.59 & 0.64  \\
    DiffScene & 0.85 & 0.87 & 0.79 & 0.84  \\
    \midrule
    \multicolumn{5}{c}{Planning-based} \\
    \noalign{\vspace{1.5pt}}
    \hdashline
    \noalign{\vspace{1.5pt}}
    TTSG (simple) & 0.22 & 0.36 & 0.28 & 0.29  \\
    TTSG (hard) & 0.74 & 0.76 & 0.66 & 0.72 \\
    \bottomrule
  \end{tabular}}
\end{table}

We evaluate the criticality of our generated scenes using SafeBench and follow the setup from DiffScene~\cite{xu2025diffscene} by selecting three representative scenario types: \textit{crossing negotiation}, \textit{red-light running}, and \textit{lane changing}, as shown in~\cref{tab:appdx_criticality}. Since the goal of this experiment is to evaluate the ability to \textit{create risk}, a higher \textit{collision rate} indicates stronger criticality. Baselines also include Learning-to-Collide (LC)~\cite{ding2020learning}, AdvSim~\cite{wang2021advsim}, and Adversarial Trajectory (AT)~\cite{zhang2022adversarial}. Following DiffScene, we evaluate each scenario across ten routes, with each route tested 10 times, resulting in 100 evaluations per scenario.

A key advantage of our framework is its flexibility in adjusting scenario criticality at test time without retraining. Unlike training-based approaches, our text-driven pipeline allows scene parameters, such as the speed of surrounding vehicles or the timing of a threat, to be modified directly through our rendering interface. To showcase this, we generate three different criticality levels per scene. Note that these variations are only applied during evaluation; in~\cref{sec:exp_gen_critical}, the scene parameters are fixed for training the ego agent.

Although our method does not achieve the highest collision rate among all baselines, it is important to note that our approach is planning-based and does not rely on adversarial training or gradient-based optimization to generate risky scenarios. Despite this, it outperforms several training-based methods, demonstrating the effectiveness of our pipeline in producing meaningful critical scenarios through structured reasoning rather than optimization. Moreover, since our method provides a reasoning-driven layout, grounded in agent intent, spatial relationships, and scene semantics, the resulting risk emerges in a natural and interpretable manner, ensuring that the generated scenarios are both plausible and actionable for downstream evaluation or training.

\section{Failure Rate}\label{sec:failure}

\begin{table}[t]
\centering
\caption{Failure rate across different LLMs.}
\label{tab:failure}
\resizebox{\linewidth}{!}{\begin{tabular}{cccc}
\toprule
\textbf{Gemma3-12B} & \textbf{Gemini-2.5-Flash} & \textbf{GPT-4o} & \textbf{Claude-Sonnet-3.5} \\
\midrule
 7\% & 1\% & 1\% & 0\% \\
\bottomrule
\end{tabular}}
\end{table}

As described in~\cref{sec:llm}, we re-submit the prompt to the LLM whenever parsing fails or the output contains unsupported options. While this mechanism may incur additional token usage and computational overhead, we demonstrate in~\cref{tab:failure} that such cases are rare for both open and closed-source models. To evaluate robustness, we reuse the same set of prompts from the diversity test, resulting in 40 runs per model. The results show that most closed-source models achieve near-perfect success rates, while open-source models still maintain high reliability, with up to 93\% success. Note that failure rates are computed across all three stages, including prompt analysis, road retrieval, and agent planning.

\section{Details for Scenarios on SafeBench}\label{sec:appdx_critical}

We conduct experiments across six different critical scenarios, three used for training the ego agent and three for evaluating scene criticality. The evaluation metrics are defined as follows: \textit{collision rate} measures how frequently collisions occur between the ego vehicle and other agents; \textit{route incompleteness} quantifies the proportion of the route left unfinished when the scenario ends; and the \textit{driving score} is a overall composite metric that incorporates collision rate, rule violation, route following, smoothness, and speed to provide a comprehensive assessment of driving performance.

\paragraph{Straight Obstacle.} In this scenario, an agent suddenly appears in front of the ego vehicle. For example, a pedestrian may cross the road as the ego vehicle approaches. In this setting, the ego agent should learn to either swerve to another lane or slow down when detecting potential risks.

\paragraph{Lane Changing.} In this scenario, the ego vehicle encounters other cars performing unexpected lane changes from different directions, without awareness of the ego vehicle's behavior. The ego agent must learn to proactively avoid these dangerous maneuvers.

\paragraph{Unprotected Left Turn.} In this scenario, an oncoming vehicle and the ego vehicle reach an intersection simultaneously, with potential trajectory overlap. Since there are no stop signs or traffic lights, the ego agent must learn how to yield or adjust its movement to avoid a collision.

\paragraph{Crossing Negotiation.} In this scenario, a vehicle approaches the intersection without traffic lights. The ego vehicle must negotiate with the other vehicle to determine the appropriate time to proceed to avoid potential collisions.

\paragraph{Red-Light Running.} In this scenario, the ego vehicle is crossing an intersection while another vehicle from a perpendicular direction runs a red light, creating a high risk of collision. The ego vehicle must anticipate and respond to these rule-violating agents.

\paragraph{Right Turn.} In this scenario, the ego vehicle is preparing to make a right turn while another vehicle from either the left or right lane also attempts to enter the same lane. The ego vehicle must remain cautious and complete the turn safely while accounting for surrounding traffic.

\section{Prompts for Ablation Study}\label{sec:appdx_abl}

We present the detailed prompts used to evaluate each pipeline design across different scenario types below.

\paragraph{Normal:}
\begin{itemize}
    \item \textit{The ego car is going straight}
    \item \textit{Three cars including the ego car are driving. The car in front go straight. The ego is turning right. The car behind the ego is turning left}
    \item \textit{A bus coming from the left road is turning left. A truck from the opposite straight is turning right. The ego car is turning right. Two cars in front of the ego car are going straight}
    \item \textit{The ego vehicle is turning left. A pedestrian on the destination suddenly block the ego}
\end{itemize}

\paragraph{Critical:}
\begin{itemize}
    \item \textit{A dangerous motorcycle on the right front is trying to turn left. The ego car is going straight}
    \item \textit{A car on the front left is trying to block the ego car. A dangerous pedestrian on the shoulder right in front of a stopped truck is crossing the road. Both the truck and the pedestrian are in the front right of the ego car}
    \item \textit{Two cars from the opposite straight is coming when the ego car is turning left}
\end{itemize}
\paragraph{Conditional}
\begin{itemize}
    \item \textit{The ego car is turning left at the intersection with no traffic light and stop sign. Three cars from the opposite straight are turning right}
    \item \textit{The ego car is going straight at the intersection with a traffic light. There are some puddles on the road}
    \item \textit{A pedestrian is crossing the road with the parallel open crosswalk and the ego car is turning left}
\end{itemize}

\section{Metrics Definition}\label{sec:metrics}

In this section, we define the metrics used to evaluate our framework.

\paragraph{Agent Accuracy.}  
Given the ground-truth agent plan containing $n$ agents $\mathbf{A}_G \triangleq \{ A^1_G, \dots, A^n_G \}$ and the predicted agent plan containing $m$ agents $\mathbf{A}_P \triangleq \{ A^1_P, \dots, A^m_P \}$, we define \textit{Agent Accuracy} (AA) as the proportion of correctly matched agents, accounting for both missing and hallucinated agents. A predicted agent can be matched to at most one ground-truth agent. Formally, let $\text{Match}(A^i_G, \mathbf{A}_P)$ denote whether a matching agent exists in the prediction set. Once a predicted agent is matched, it is removed from consideration in subsequent matches. The accuracy is computed as:
\begin{equation*}
    \text{AA} = \frac{1}{\max(m, n)} \sum_{i=1}^{n} \1_\mathrm{\left\{ \exists A^j_P \in \mathbf{A}_P \text{ such that } A^j_P \equiv A^i_G \right\}},
\end{equation*}
where $\equiv$ denotes a match based on key attributes (\eg, type, position, and action), and each matched $A^j_P$ is removed from $\mathbf{A}_P$ to prevent multiple assignments. The use of $\max(m, n)$ penalizes both over-generation (hallucinated agents) and under-generation (missed agents).

\paragraph{Road Accuracy.}  
Given a ground-truth road specification with $n$ properties $R_G \triangleq \{ r^1_G, \dots, r^n_G \}$ and a predicted road with $m$ properties $R_P \triangleq \{ r^1_P, \dots, r^m_P \}$, we define \textit{Road Accuracy} (RA) as the proportion of correctly predicted properties. Each property may refer to a structural or semantic attribute, such as the number of lanes, presence of specific signals, or existence of certain static objects. Road accuracy is computed as:
\begin{equation*}
    \text{RA} = \frac{1}{\max(m, n)} \sum_{i=1}^n \1_\mathrm{{\left\{ \exists r^j_P \in R_P \text{ such that } r^j_P = r^i_G \right\}}},
\end{equation*}
where a predicted property is considered correct if it matches any unmatched ground-truth property. As with agent accuracy, we normalize by $\max(m, n)$ to penalize both missing and extra (hallucinated) properties, ensuring balanced evaluation across under- and over-prediction cases.

\paragraph{Agent Diversity.}  
Given a set of $n$ generated agents $\mathbf{A}_p = \{ A_p^1, \dots, A_p^n \}$ derived from the same input prompt $p$, we define \textit{Agent Diversity (AD)} as the proportion of unique agent configurations, reflecting variation in scene composition. Agent Diversity is computed as:
\begin{equation*}
    \text{AD} = \frac{|\text{set}(\mathbf{A}_p)|}{n},
\end{equation*}
where $\text{set}(\mathbf{A}_p)$ denotes the number of unique agent instances, determined based on combinations of agent type, action, and relative position.

\paragraph{Road Diversity.}  
Given a set of $n$ generated roads $\mathbf{R}_p = \{ R_p^1, \dots, R_p^n \}$ corresponding to the same input prompt $p$, we define \textit{Road Diversity (RD)} as the proportion of unique roads selected for generation. The uniqueness is determined based on the road ID assigned in the simulator. The road diversity is computed as:
\begin{equation*}
    \text{RD} = \frac{|\text{set}(\mathbf{R}_p)|}{n},
\end{equation*}
where $\text{set}(\mathbf{R}_p)$ returns the number of distinct road IDs in $\mathbf{R}_p$.

\section{Limitation}

As our framework relies on LLMs, its planning performance can be affected by limitations in language understanding. Despite our reasoning stage, the LLM may misinterpret prompts. For instance, failing to recognize ``{\tt \small A car in front does not move for an infinite amount of time}'' as a \textit{blocking} action. Hallucinations are another issue; the LLM sometimes inserts elements like traffic lights even when not specified. These issues may be mitigated by using more capable reasoning models (\eg, OpenAI-o1~\cite{jaech2024openai}) or fine-tuning open-source LLMs with expert guidance~\cite{sun-etal-2024-aligning}.

\paragraph{Limit Cases}
Another source of failure arises from missing signals and objects within CARLA. Since our framework relies on the built-in map, complex combinations of signals and objects, and novel conditions cannot be retrieved. For instance, identifying a road, such as allowing all directions but lacking traffic lights or stop signs, is impossible. Similarly, unsupported conditions. For example, requesting a speed limit of ``200'' cannot be fulfilled. To address this, we plan to support custom OpenDRIVE code generation, enabling flexible scene customization.
\end{appendices}


\end{document}